# Reinforcement Learning in Partially Observable Markov Decision Processes using Hybrid Probabilistic Logic Programs

Emad Saad
Department of Computer Science
Gulf University for Science and Technology
Mishref, Kuwait
saad.e@gust.edu.kw

*Abstract*—We present a probabilistic logic programming framework to reinforcement learning, by integrating reinforcement learning, in POMDP environments, with normal hybrid probabilistic logic programs with probabilistic answer set semantics, that is capable of representing domain-specific knowledge. We formally prove the correctness of our approach. We show that the complexity of finding a policy for a reinforcement learning problem in our approach is NP-complete. In addition, we show that any reinforcement learning problem can be encoded as a classical logic program with answer set semantics. We also show that a reinforcement learning problem can be encoded as a SAT problem. We present a new high level action description language that allows the factored representation of POMDP. Moreover, we modify the original model of POMDP so that it be able to distinguish between knowledge producing actions and actions that change the environment.

## I. INTRODUCTION

Reinforcement learning is the problem of learning to act by trial and error interaction in dynamic environments. Reinforcement learning problems can be represented as Markov Decision Processes (MDP), under the assumption that accurate and complete model of the environment is known. This assumption requires the agent to have perfect sensing and observation abilities.

However, complete and perfect observability is unrealistic for many real-world reinforcement learning applications, although necessary for learning optimal policies in MDP environments. Therefore, different model is needed to represent and solve reinforcement learning problems with partial observability. This model is Partially Observable Markov Decision Processes (POMDP). Similar to MDP, POMDP requires the model of the environment to be known, however states of the world are not completely known. Consequently, the agent performs actions to make observations about the states of the worlds. These observations can be noisy due to imperfect agent's sensors. Similar to MDP, dynamic programming methods, by value iteration, has been used to learn the optimal policy for a reinforcement learning problem in POMDP environment.

A logical framework to reinforcement learning in MDP environment has been developed in [30], which relies on techniques from probabilistic reasoning and knowledge representation by normal hybrid probabilistic logic programs [34]. The normal hybrid probabilistic logic programs framework of [30] has been proposed upon observing that dynamic programming methods to reinforcement learning in general and value iteration in particular are incapable of exploiting domain-specific knowledge of the reinforcement learning problem domains to improve the efficiency of finding the optimal policy. In addition, these dynamic programming methods use primitive representation of states and actions as this representation does not capture the relationship between states [22] and makes it difficult to represent domain-specific knowledge. However, using richer knowledge representation frameworks for MDP and POMDP allow efficiently finding optimal policies in more complex stochastic domains and lead to develop methods to find optimal policies with larger domains sizes [22].

The choice of normal hybrid probabilistic logic programs (NHPLP) to solve reinforcement learning problems in MDP environment is based on that; NHPLP is nonmonotonic, therefore more suitable for knowledge representation and reasoning under uncertainty; NHPLP subsumes classical normal logic programs with classical answer set semantics [7], a rich knowl-edge representation and reasoning framework, and inherits its knowledge representation and reasoning capabilities including the ability to represent and reason about domain-specific knowledge; NHPLP has been shown applicable to a variety of fundamental probabilistic reasoning problems including probabilistic planning [28], contingent probabilistic planning [31], the most probable explanation in belief networks, the most likely trajectory in probabilistic planning, and Bayesian reasoning [29].

In this view, we integrate reinforcement learning in POMDP environment with NHPLP, providing a logical framework that overcomes the representational limitations of dynamic programming method to reinforcement learning in POMDP and is capable of representing its domain-specific knowledge. In addition, the proposed framework extends the logical framework of reinforcement learning in MDP of [30] with partial observability. We show that any reinforcement learning problem in POMDP environment can be encoded as a SAT problem. The importance of that is reinforcement learning problems in POMDP environment can be now solved as SAT problems.

## II. SYNTAX AND SEMANTICS OF NHPLP

We introduce a class of NHPLP [34], namely NHPLP$_{\mathcal{PO}}$ that is sufficient to represent POMDP.

### A. The Language of NHPLP$_{\mathcal{PO}}$

Let $\mathcal{L}$ be a first-order language with finitely many predicate symbols, constants, and infinitely many variables. The Herbrand base of $\mathcal{L}$ is denoted by $\mathcal{B}_{\mathcal{L}}$. Probabilities are assigned to atoms in $\mathcal{B}_{\mathcal{L}}$ as values from $[0,1]$. An *annotation*, $\mu$, is either a constant in $[0,1]$, a variable (*annotation variable*) ranging over $[0,1]$, or $f(\mu_1, \ldots, \mu_n)$ (called *annotation function*) where $f$ is a representation of a computable total function $f : ([0,1])^n \to [0,1]$ and $\mu_1, \ldots, \mu_n$ are annotations. Let $a_1, a_2 \in [0,1]$. Then we say that $a_1 \leq_t a_2$ iff $a_1 \leq a_2$. A normal probabilistic logic program (*np-program*) in NHPP$_{\mathcal{PO}}$ is a pair $P = \langle R, \tau \rangle$, where $R$ is a finite set of normal probabilistic rules (np-rules) and $\tau$ is a mapping $\tau : \mathcal{B}_{\mathcal{L}} \to S_{disj}$, where $S_{disj}$ is a set of disjunctive probabilistic strategies (p-strategies) whose composition functions $c$, are mappings $c : [0,1] \times [0,1] \to [0,1]$. A composition function of a disjunctive p-strategy returns the probability of a disjunction of two events given the probability values of its components An np-rule is an expression of the form
$$A : \mu \leftarrow A_1 : \mu_1, \ldots, A_n : \mu_n,$$
$$not\ (B_1 : \mu_{n+1}), \ldots, not\ (B_m : \mu_{n+m})$$
where $A, A_1, \ldots, A_n, B_1, \ldots, B_m$ are atoms and $\mu, \mu_i\ (1 \leq i \leq m+n)$ are annotations. Intuitively, the meaning of an np-rule is that if for each $A_i : \mu_i$, the probability of $A_i$ is at least $\mu_i$ (w.r.t. $\leq_t$) and for each $not\ (B_j : \mu_j)$, it is not *believable* that the probability of $B_j$ is at least $\mu_j$, then the probability of $A$ is $\mu$. The mapping $\tau$ associates to each atom $A$ a disjunctive p-strategy that will be employed to combine the probability values obtained from different np-rules having $A$ in their heads. An np-program is ground if no variables appear in any of its p-rules.

### B. Probabilistic Answer Set Semantics of NHPLP$_{\mathcal{PO}}$

A probabilistic interpretation (p-interpretation), $h$, is a mapping from $\mathcal{B}_{\mathcal{L}}$ to $[0,1]$. Let $P = \langle R, \tau \rangle$ be a ground np-program, $h$ be a p-interpretation, and $r$ be an np-rule as above. Then, we say
- $h$ satisfies $A_i : \mu_i$ iff $\mu_i \leq_t h(A_i)$.
- $h$ satisfies $not\ (B_j : \beta_j)$ iff $\beta_j \not\leq_t h(B_j)$.
- $h$ satisfies $Body \equiv A_1 : \mu_1, \ldots, A_n : \mu_n, not\ (B_1 : \beta_1), \ldots, not\ (B_m : \beta_m)$ iff $\forall (1 \leq i \leq n), h$ satisfies $A_i : \mu_i$ and $\forall (1 \leq j \leq m), h$ satisfies $not\ (B_j : \beta_j)$.
- $h$ satisfies $A : \mu \leftarrow Body$ iff $h$ satisfies $A : \mu$ or $h$ does not satisfy $Body$.
- $h$ satisfies $P$ iff $h$ satisfies every np-rule in $R$ and for every atom $A \in \mathcal{B}_{\mathcal{L}}$, we have $c_{\tau(A)}\{\!\{\mu | A : \mu \leftarrow Body \in R$ such that $h \models Body\}\!\} \leq_t h(A)$.

The probabilistic reduct $P^h$ of $h$ w.r.t. $h$ is an np-program without negation, $P^h = \langle R^h, \tau \rangle$, where:
$$A : \mu \leftarrow A_1 : \mu_1, \ldots, A_n : \mu_n \in R^h$$
iff
$$A : \mu \leftarrow A_1 : \mu_1, \ldots, A_n : \mu_n,$$
$$not\ (B_1 : \beta_1), \ldots, not\ (B_m : \beta_m) \in R$$
and $\forall (1 \leq j \leq m), \beta_j \not\leq_t h(B_j)$. A probabilistic model (*p-model*) of an np-program $P$ is a p-interpretation of $P$ that satisfies $P$. We say that a p-interpretation $h$ of $P$ is a probabilistic answer set of $P$ if $h$ is the minimal p-model of the probabilistic reduct, $P^h$, of $P$ w.r.t. $h$.

## III. PARTIALLY OBSERVABLE MARKOV DECISION PROCESSES

We review finite-horizon POMDP [12] with stationary transition functions, stationary bounded reward functions, and stationary policies.

### A. POMDP Definition

POMDP is a tuple of the form $\mathbf{M} = \langle S, S_0, A, T, \lambda, \mathcal{R}, \Omega, O \rangle$ where: $S$ is a finite set of states; $S_0$ is the initial state distribution; $A$ is a finite set of stochastic actions; $T$ is stationary transition function $T : S \times A \times S \to [0,1]$, where for any $s \in S$ and $a \in A$, $\sum_{s' \in S} T(s, a, s') = 1$; $\lambda \in [0,1)$ is the discount factor; $\mathcal{R} : S \times A \times S \to \mathbb{R}$ is a stationary bounded reward function; $\Omega$ is a finite set of observations that the agent observes in the environment; and $O$ is observation function $O : S \times A \times \Omega \to [0,1]$, where for any $s \in S$ and $a \in A$ where $\sum_{o \in \Omega} O(s, a, o) = 1$. A stationary policy is a mapping from states to actions of the form $\pi : S \to A$. The value function of a policy $\pi$ with respect to an initial state $s_0 \in S_o$, with finite horizon of $n$ steps remaining, $V_n^\pi(s_0)$, is calculated by
$$V_n^\pi(s_0) = \sum_{s_1 \in S} T(s_0, \pi(s_0), s_1) \sum_{o_i \in \Omega} O(s_1, \pi(s_0), o_i)$$
$$\left[ \mathcal{R}(s_0, \pi(s_0), s_1) + \lambda\ V_{n-1}^\pi(s_1) \right]$$
which determines the expected sum of discounted rewards resulting from executing the policy $\pi$ starting from $s_0$. Because of the agent is unable to completely observe the states of the world and with reliability, it keeps what is called a belief state. An agent's belief state is a probability distribution over the possible world states the agent may think it is in. Therefore, an action causes a transition from a belief state to another belief state. Given $b$ is a believe state and $a$ is an action, then executing $a$ in the belief state $b$ results a new belief state $b'$, where the probability of a state, $s'$, in $b'$ and the value function of executing a policy $\pi$ in $b$ are given by:
$$b'(s') = \frac{O(s', a, o) \sum_{s \in S} T(s, a, s') b(s)}{Pr(o|a, b)}$$
$$V_n^\pi(b) = \sum_{s \in S} b(s) V_n^\pi(s).$$

The optimal policy over the agent's belief states can constructed from the optimal value function over the agent's belief states which is given by $V_n^*(b) = \max_\pi V_n^\pi(b)$.

*B. Discussion*

The original model of POMDP does not distinguish between knowledge producing (sensing) actions and actions that affects and change the environment (non-sensing actions). This means that it treats sensing and non-sensing actions equally in the sense that, like non-sensing actions, a sensing action affects and changes the environment as well as producing knowledge resulting from observing the environment. However, [36] proved that sensing actions produce knowledge (make observations) and does not change the state of the world. Therefore, actions that change the state of the world are different from the knowledge producing actions. In addition, the value function described above makes the agent observing the environment at every step of its life with each action it takes. However, this is not necessary to be always the case, since it is possible for the agent to start with observing the environment then performing a sequence of actions, or the agent could start with performing a sequence of actions then observing the environment. To overcome these limitations, we define the value function of n-step finite horizon POMDP with respect to an initial state $s_0 \in S_o$ as:

- if $\pi(s_0)$ is a non-sensing action

$$V_n^\pi(s_0) = \sum_{s_1 \in S} T(s_0, \pi(s_0), s_1)[\mathcal{R}(s_0, \pi(s_0), s_1) + \lambda\, V_{n-1}^\pi(s_1)]$$

- if $\pi(s_0)$ is sensing action

$$V_n^\pi(s_0) = \sum_{s_1 \in S} O(s_0, \pi(s_0), s_1)[\mathcal{R}(s_0, \pi(s_0), s_1) + \lambda\, V_{n-1}^\pi(s_1)]$$

where $O(s_0, \pi(s_0), s_1)$ is the probability of observing the state $s_1$, where for some $o \in \Omega$, $o$ is observed in $s_1$. Notice that $O$ is treated as a mapping $O : S \times A \times S \to [0, 1]$, where $A$ is the set of sensing actions. For any $s \in S$ and $a \in A$, $O(s, a, .)$ is the probability distribution over states resulting from executing $a$ in $s$, such that $\sum_{s' \in S} O(s, a, s') = 1$. As in the original model of POMDP, $T$ is a mapping $T : S \times A \times S \to [0, 1]$, where $A$ is the set of non-sensing actions. Extension to infinite horizon POMDP can be achieved in a similar manner. This definition of POMDP distinguishes between knowledge producing actions and actions that change the environment. In this view, the optimal policy $V_n^*$ is given by: $V_n^*(s_0) = \max_\pi V_n^\pi(s_0)$

## IV. $\mathcal{A}_{\mathcal{PO}}$ AN ACTION LANGUAGE FOR POMDP

We introduce an action language for POMDP, $\mathcal{A}_{\mathcal{PO}}$. The proposed action language extends both the action language, $\mathcal{A}_{MD}$, [30] for representing and reasoning about MDP, and the action language, $\mathcal{P}$, [31] for representing and reasoning about imperfect sensing actions with probabilistic outcomes. An action theory in $\mathcal{A}_{\mathcal{PO}}$ is capable of representing the initial state distribution, the executability conditions of actions, the discount factor, the reward received from executing actions in states, and makes it clear the distinction between sensing and non-sensing actions.

*A. Language syntax*

A fluent is a predicate, which may contain variables. Given that $\mathcal{F}$ is a set of fluents and $\mathcal{A}$ is a set of actions that can contain variables, a fluent literal is either a fluent $f \in \mathcal{F}$ or $\neg f$. A conjunction of fluent literals of the form $l_1 \wedge \ldots \wedge l_n$ is conjunctive fluent formula, where $l_1, \ldots, l_n$ are fluent literals. Sometimes we abuse the notation and refer to a conjunctive fluent formula as a set of fluent literals ($\emptyset$ denotes $true$). An action theory, $\mathbf{PT}$, in $\mathcal{A}_{PO}$ is a tuple $\mathbf{PT} = \langle S_0, \mathcal{D}, \lambda \rangle$, where $S_0$ is a proposition of the form (1), $\mathcal{D}$ is a set of propositions from (2-4), and $0 \leq \lambda < 1$ is a discount factor as follows:

$$\mathbf{initially}\ \{\ \psi_i\ :\ p_i,\quad 1 \leq i \leq n \quad (1)$$

$$\mathbf{executable}\ a\ \mathbf{if}\ \psi \quad (2)$$

$$a\ \mathbf{causes}\ \{\ \phi_i\ :\ p_i\ :\ r_i\ \mathbf{if}\ \psi_i,\quad 1 \leq i \leq n \quad (3)$$

$$a\ \mathbf{observes}\ \{\ o_i\ :\ p_i\ :\ r_i\ \mathbf{sensing}\ \psi_i,\quad 1 \leq i \leq n \quad (4)$$

where $\psi, \psi_i, \phi_i, o_i, (1 \leq i \leq n)$ are conjunctive fluent formulas, $a \in \mathcal{A}$, and $p_i \in [0,1]$. The set of all ground $\psi_i$ and $o_i$ must be exhaustive and mutually exclusive.

The *initial agent's belief state*—a probability distribution over the possible initial states, is represented by (1), that says each possible initial state $\psi_i$ holds with probability $p_i$. *Executability condition* is represented by (2). A non-sensing action, $a$, is represented by (3), which says that for each $1 \leq i \leq n$, $a$ causes $\phi_i$ to hold with probability $p_i$ and reward $r_i$ is received in a successor state to a state in which $a$ is executed and $\psi_i$ holds. A sensing action, $a$, is represented by (4), which says that for each $1 < i \leq n$, whenever a correlated $\psi_i$ is known to be true, $a$ causes any of $o_i$ to be known true with probability $p_i$ and reward $r_i$ is received in a successor state to a state in which $a$ is executed, where the literals in $\psi_i$ determine what the agent is observing and literals in $o_i$ determine what the sensor reports on. Similar to [4], when a property of the world cannot be directly sensed by the sensor, another correlated property of the world, that can be sensed by the sensor, can be used instead. An action theory is ground if it does not contain any variables.

In the sequel, we represent an action $a$ in (3) as a set of the form $a = \{a_1, \ldots, a_n\}$, where each $a_i$ corresponds to $\phi_i$, $p_i$, $r_i$, and $\psi_i$. For each $1 < i \leq n$, (3) can be represented as $a_i$ **causes** $\phi_i\ :\ p_i\ :\ r_i$ **if** $\psi_i$. Similarly, (4) can be represented as $a_i$ **observes** $o_i\ :\ p_i\ :\ r_i$ **sensing** $\psi_i$.

*Example 1:* Consider the tiger domain from [20]. A tiger is behind left ($tl$) or tight ($\neg tl$) door with equal probability 0.5. If left door is opened and ($tl$), punishment of -100 is received, but a reward of 10 is received if ($\neg tl$) and the other way around. The sensing action *listen* used for hearing the tiger behind left door ($htl$), a correlated property to $tl$. But, the agent's hearing is not perfect and costs -1. If the agent listens to ($htl$), then it reports $tl$ with 0.85 and erroneously reports $\neg tl$ with 0.15. Similarly for listening to the right door. This is represented by the action theory $\mathbf{PT} = \langle S_0, \mathcal{D}, \lambda \rangle$, where

$$S_0 = \mathbf{initially} \begin{cases} \{tl, htl\} & :\ 0.5 \\ \{\neg tl, \neg htl\} & :\ 0.5 \end{cases}$$

**executable** $AC$ **if** $\emptyset$, where $AC \in \{openL, openR, listen\}$

$$openL \textbf{ causes } \begin{cases} \{tl\} & : & 1 & : & -100 & \textbf{if} & \{tl\} \\ \{\neg tl\} & : & 1 & : & 10 & \textbf{if} & \{\neg tl\} \end{cases}$$

$$openR \textbf{ causes } \begin{cases} \{\neg tl\} & : & 1 & : & -100 & \textbf{if} & \{\neg tl\} \\ \{tl\} & : & 1 & : & 10 & \textbf{if} & \{tl\} \end{cases}$$

$$listen \textbf{ observes } \begin{cases} \{tl\} & : & 0.85 & : & -1 & \textbf{sensing} & \{htl\} \\ \{\neg tl\} & : & 0.15 & : & -1 & \textbf{sensing} & \{htl\} \\ \{\neg tl\} & : & 0.85 & : & -1 & \textbf{sensing} & \{\neg htl \\ \{tl\} & : & 0.15 & : & -1 & \textbf{sensing} & \{\neg htl \end{cases}$$

*B. Semantics*

A set of ground literals $\phi$ is consistent if it does not contain a pair of complementary literals. If a literal $l$ belongs to $\phi$, then we say $l$ is true in $\phi$, and $l$ is false in $\phi$ if $\neg l$ is in $\phi$. A set of literals $\sigma$ is true in $\phi$ if $\sigma$ is contained in $\phi$. A state $s$ is a complete and consistent set of literals that describes the world at a certain time point.

*Definition 2:* Let $\textbf{PT} = \langle S_0, \mathcal{D}, \lambda \rangle$ be a ground action theory in $\mathcal{A}_{\mathcal{PO}}$, $s$ be a state, $a_i \textbf{ causes } \phi_i : p_i : r_i \textbf{ if } \psi_i$ $(1 \leq i \leq n)$ be in $\mathcal{D}$, and $a = \{a_1, \ldots, a_n\}$ be an action, where each $a_i$ corresponds to $\phi_i, p_i, r_i$, and $\psi_i$ for $1 \leq i \leq n$ (similarly for $a_i \textbf{ observes } \phi_i : p_i : r_i \textbf{ sensing } \psi_i$). Then, the state resulting from executing $a$ in $s$ $\Phi(a,s)$ is:
- $l \in \Phi(a,s)$ and $\neg l \notin \Phi(a,s)$ iff $l \in \phi_i$ and $\psi_i \subseteq s$.
- $\neg l \in \Phi(a,s)$ and $l \notin \Phi(a,s)$ iff $\neg l \in \phi_i$ and $\psi_i \subseteq s$.
- Else $l \in \Phi(a,s)$ iff $l \in s$ and $\neg l \in \Phi(a,s)$ iff $\neg l \in s$.

*Definition 2:* Let $s$ be a state, and $a_i \textbf{ causes } \phi_i : p_i : r_i \textbf{ if } \psi_i$ (similarly $a_i' \textbf{ observes } o_i : p_i' : r_i' \textbf{ sensing } \psi_i$) $(1 \leq i \leq n)$ be in propositions. Then, the transition probability distribution after executing $a$ ($a'$) in $s$ is given by
$$T(s,a,s') = \begin{cases} p_i & if s' = \Phi(a_i, s) \\ 0 & otherwise \end{cases}$$
$$O(s,a',s') = \begin{cases} p_i' & if s' = \Phi(a_i', s) \\ 0 & otherwise \end{cases}$$
The reward received in a state $s'$ after executing $a$ ($a'$) in $s$ is $\mathcal{R}(s,a,s') = r_i$ if $s' = \Phi(a_i, s)$, $\mathcal{R}(s,a',s') = r_i'$ if $s' = \Phi(a_i', s)$, otherwise $\mathcal{R}(s,a,s') = \mathcal{R}(s,a',s') = 0$.

*Definition 3:* Let $s_0$ be an initial state, $s, s'$ be states, and $\pi$ be a policy in $\textbf{PT}$. Then, the value function of n-step remaining, $V_n^\pi$, of $\pi$ is given by:
- if $\pi(s_0)$ is a non-sensing action and $X = T(s_0, \pi(s_0), s_1)$
$V_n^\pi(s_0) = \sum_{s_1 \in S} X \left[ \mathcal{R}(s_0, \pi(s_0), s_1) + \lambda\, V_{n-1}^\pi(s_1) \right]$
- if $\pi(s_0)$ is sensing action and $Y = O(s_0, \pi(s_0), s_1)$ $V_n^\pi(s_0)$
$V_0^\pi(s_n) = \mathcal{R}(s_{n-1}, \pi(s_{n-1}), s_n)$.

Executing sensing or non-sensing action, $\pi(s)$, in $s$ causes a transition to a set of states, $\sigma = \{s_1', s_2', \ldots, s_m'\}$. Let $\pi(\sigma)$ denotes the set of actions $\pi(s_1'), \pi(s_2'), \ldots, \pi(s_m')$ executed in the states $s_1', s_2', \ldots, s_m'$ respectively. Notice that if $\pi(\sigma)$ is a singleton, i.e., the same action is executed in every state in $\sigma$, then this corresponds to executing an action in a belief state $\sigma = \{s_1', s_2', \ldots, s_m'\}$. Since executing $\pi(\sigma)$ in $\sigma$ produces another set of states $\sigma'$, then executing $\pi(\sigma)$ causes a transition from a belief state to another belief state.

For finite horizon POMDP, a policy $\pi : S \to \mathcal{A}$ can be represented as a set of ordered pairs, starting from the initial belief state $\sigma_0$ (the set of initial states in $S_0$), as $\pi = \{(\sigma_0, \pi(\sigma_0)), (\sigma_1, \pi(\sigma_1)), \ldots, (\sigma_{n-1}, \pi(\sigma_{n-1}))\}$ where for $1 \leq i \leq n$, $\sigma_i$ represents a belief state (a set of states) resulting from executing $\pi(\sigma_{i-1})$ in $\sigma_{i-1}$. This set representation of finite horizon policies in POMDP leads to view a policy as a set of trajectories, where each trajectory is $j(n) \equiv s_0, \pi(s_0), s_1, \pi(s_1), \ldots, s_{n-1}, \pi(s_{n-1}), s_n$ where $s_0$ is an initial state in $S_0$ and for all $1 \leq i \leq n$, $s_i \in \sigma_i$ and $\pi(s_i) \in \pi(\sigma_i)$, such that for any $1 \leq i \leq n$, $s_i = \Phi(s_{i-1}, \pi(s_{i-1}))$. Let $\pi$ be a policy for a finite horizon POMDP and $T_\pi$ be the set of trajectories representation of $\pi$, given the trajectory view of $\pi$, the value function of $\pi$ can be now described as:
$$V_n^\pi(s_0) = \sum_{j(n) \in T_\pi} \left[ \sum_{t=0}^{n-1} \lambda^t \left[ \prod_{i=0}^{t} X(s_i, \pi(s_i), s_{i+1}) \right] \mathcal{R}(s_t, \pi(s_t), \cdot \right]$$
where
$$X(s_i, \pi(s_i), s_{i+1}) = \begin{cases} T(s_i, \pi(s_i), s_{i+1}), \pi(s_i) \text{ is nonsensing} \\ O(s_i, \pi(s_i), s_{i+1}), \pi(s_i) \text{ is sensing} \end{cases}$$
Thus, the optimal policy $V_n^*$, the maximum value function among all policies, is given by $V_n^*(s_0) = \max_\pi V_n^\pi(s_0)$.

V. REINFORCEMENT LEARNING IN NHPLP$_{\mathcal{PO}}$

This section uses NHPLP$_{\mathcal{PO}}$ to solve reinforcement learning problems, by encoding an action theory, $\textbf{PT}$ in $\mathcal{A}_{\mathcal{PO}}$, into an np-program, $\Pi_{\textbf{PT}}$. The probabilistic answer sets of $\Pi_{\textbf{PT}}$ correspond to valid trajectories in $\textbf{PT}$ with associated value function. The np-program encoding of an action theory in $\mathcal{A}_{\mathcal{PO}}$ follows related encoding described in [30], [31], [37]. We assume that the length of the optimal policy that we are looking for is known and finite. We use the following predicates: $holds(L, T)$ for literal $L$ holds at time moment $T$, $occ(A, T)$ for action $A$ executes at time $T$, $state(T)$ for a state of the world at time $T$, $reward(T, r)$ for the reward received at time $T$ is $r$, $value(T, V)$ for the value function of a state at time $T$ is $V$, and $factor(\lambda)$ for the discount factor $\lambda$. If an atom appears in an np-rule in $R$ with no annotation it is assumed to be associated with the annotation 1. We use $p(\psi)$ to denote $p(l_1), \ldots, p(l_n)$ for $p$ is a predicate and $\psi = \{l_1, \ldots, l_n\}$. Let $\Pi_{\textbf{PT}} = \langle R, \tau \rangle$ be the np-program encoding $\textbf{PT} = \langle S_0, \mathcal{D}, \lambda \rangle$, where $R$ is the set of the following np-rules.
- Each action $a = \{a_1, \ldots, a_n\} \in \mathcal{A}$, is encoded as
for all $1 \leq i \leq n$, $action(a_i) \leftarrow$   (6)
Each fluent $f \in \mathcal{F}$ is encoded as a fact of the form
$fluent(f) \leftarrow$. Fluent literals are encoded as
$literal(F) \leftarrow fluent(F)$   (7)
$literal(\neg F) \leftarrow fluent(F)$   (8)
To specify that fluents $F$ and $\neg F$ are contrary literals, we use the following np-rules.
$contrary(F, \neg F) \leftarrow fluent(F)$   (9)
$contrary(\neg F, F) \leftarrow fluent(F)$   (10)

- The initial belief state **initially** $\{\psi_i : p_i, 1 \leq i \leq n\}$ is represented in $R$ as follows. Let $s_1, s_2, \ldots, s_n$ be the set of possible initial states, where for each $1 \leq i \leq n$, $s_i = \{l_1^i, \ldots, l_m^i\}$, and the initial probability distribution be $Pr(s_i) = p_i$. Moreover, let $s = s_1 \cup s_2 \cup \ldots \cup s_n$, $s' = s_1 \cap s_2 \cap \ldots \cap s_n$, $\widehat{s} = s - s'$. Let $s^{report} = \{l \mid l \in s_i$ and $l$ is a sensor report literal $\}$ be the set of all sensor report literals in all $s_i$. Let $s'' = \{l \mid l \in (\widehat{s} - s^{report}) \vee \neg l \in (\widehat{s} - s^{report})\}$. Intuitively, $s''$ is the same as $\widehat{s}$ after excluding the set of sensor report literals $s^{report}$ from $\widehat{s}$. Let $s^{sense}$ be the set of all pairs $(\delta_i, \gamma_i)$, where $\delta_i$ and $\gamma_i$ are sets of literals contained in $s_i$, such that $\delta_i$ is the set of sensor reading literals and $\gamma_i$ is the set of sensor report literals appearing in $s_i$. The set of all possible initial states are generated as follows: for each $l \in s'$, we include in $R$
$$holds(l, 0) \leftarrow \quad (11)$$
which represents a fact that holds in every possible initial state. It says that the literal $l$ holds at time moment 0. In addition, for each $l \in s''$, $R$ includes
$$holds(l, 0) \leftarrow not\ holds(\neg l, 0) \quad (12)$$
$$holds(\neg l, 0) \leftarrow not\ holds(l, 0) \quad (13)$$
These np-rules say $l$ (similarly $\neg l$) holds at time moment 0, if $\neg l$ (similarly $l$) does not hold at the time moment 0. For each $(\delta, \gamma) \in \psi^{sense}$, let $\gamma = \{l_1, \ldots, l_m\}$, then for each $1 \leq i \leq m$, $R$ includes
$$holds(l_i, 0) \leftarrow holds(\delta, 0) \quad (14)$$
The initial probability distribution over the initial states is encoded as follows, which says that the probability of a state at time 0 is $p_i$, if $l_1^i, \ldots, l_m^i$ hold at the time 0.
$$state(0) : p_i \leftarrow holds(l_1^i, 0), \ldots, holds(l_m^i, 0) \quad (15)$$

- Each executability condition of an action of the form (2) is encoded for each $1 < i < n$ as
$$exec(a_i, T) \leftarrow holds(\psi, T) \quad (16)$$

- For each non-sensing action proposition $a_i$ **causes** $\phi_i : p_i : r_i$ **if** $\psi_i, 1 \leq i \leq n$, in $\mathcal{D}$, let $\phi_i = \{l_i^1, \ldots, l_i^m\}$. Then, $\forall (1 \leq j \leq m)$, $R$ includes
$$holds(l_i^j, T+1) \leftarrow occ(a_i, T), exec(a_i, T),$$
$$holds(\psi_i, T) \quad (17)$$
If $a$ occurs at time $T$ and $\psi_i$ holds at the same time moment, then the $l_i^j$ holds at the time $T+1$. Then, we have
$$state(T+1) : p_i \times U \leftarrow state(T) : U, occ(a_i, T),$$
$$exec(a_i, T), holds(\psi_i, T), holds(\phi_i, T+1) \quad (18)$$
where $U$ is an annotation variable ranging over $[0,1]$ acts as a place holder. This np-rule states that if $\psi_i$ holds in a state at time $T$, whose probability is $U$, and in which $a$ is executable, then the probability of a successor state at time $T+1$ is $p_i \times U$, in which $\phi_i$ holds.

- For each sensing action proposition $a_i$ **observes** $o_i : p_i : r_i$ **sensing** $\psi_i, 1 \leq i \leq n$, in $\mathcal{D}$, let $o_i = \{l_i^1, \ldots, l_i^m\}$ and $\psi_i = \{l_i'^1, \ldots, l_i'^m\}$. Then, $\forall (1 \leq j \leq m)$, $R$ includes
$$observed(l_i'^j, T) \leftarrow occ(a_i, T), exec(a_i, T),$$
$$holds(\psi_i, T) \quad (19)$$
$$holds(l_i^j, T+1) \leftarrow occ(a_i, T), exec(a_i, T),$$
$$observed(\psi_i, T) \quad (20)$$
where first p-rule says that executing the sensing action $a$ at time $T$ in which $\psi_i$ holds causes $\psi_i$ to be observed to be known true at the same moment $T$, and second p-rule states that if $a$ occurs at time $T$ and the literals in $\psi_i$ are observed to be known true at the same moment, then the literals $l_i^j \in o_i$ are known to hold at the time moment $T+1$.
$$state(T+1) : p_i \times U \leftarrow state(T) : U, occ(a_i, T),$$
$$exec(a_i, T), observed(\psi_i, T), holds(o_i, T+1) \quad (21)$$
The above np-rule says that the probability of a state at time $T+1$ is $p_i \times U$ if $o_i$ become known true at the same moment, after executing $a$ in a state at time $T$, whose probability is $U$, in which the literals in $\psi_i$ are observed true.

- The reward $r_i$ received at time $T+1$ after executing $a$ in a state at time $T$ is encoded as
$$reward(r_i, T+1) \leftarrow occ(a_i, T), exec(a_i, T) \quad (22)$$

- The value function $T+1$ steps away from the initial state given the value function $T$ steps away from the initial state in a given episode is encoded as

-- if $a$ is a non-sensing action
$$value(V + \lambda^T * U * r_i, T+1) \leftarrow value(V, T),$$
$$factor(\lambda), state(T+1) : U, reward(r_i, T+1),$$
$$occ(a_i, T), exec(a_i, T), holds(\psi_i, T),$$
$$holds(\phi_i, T+1) \quad (23)$$

-- if $a$ is a sensing action
$$value(V + \lambda^T * U * r_i, T+1) \leftarrow value(V, T),$$
$$factor(\lambda), state(T+1) : U, reward(r_i, T+1),$$
$$occ(a_i, T), exec(a_i, T), observed(\psi_i, T),$$
$$holds(o_i, T+1) \quad (24)$$
where the variables $V \in \mathbb{R}$, $\lambda \in [0,1)$, $U \in [0,1]$, and $factor(\lambda)$ is a fact in $R$. These np-rules state that the value function at time $T+1$ is equal to the value function at time $T$ added to the product of the reward $r_i$ received in a state at time $T+1$ and the probability of a state at time $T+1$ discounted by $\lambda^T$.

- The following np-rule asserts that a literal $L$ holds at $T+1$ if it holds at $T$ and its contrary does not hold at $T+1$.
$$holds(L, T+1) \leftarrow holds(L, T),$$
$$not\ holds(L', T+1), contrary(L, L') \quad (25)$$

- The literal, $A$, and its negation, $\neg A$, cannot hold at the same time, where $inconsistent$ is a literal that does not appear in **PT**.

$$inconsistent \leftarrow not\ inconsistent, holds(A, T),$$
$$holds(\neg A, T) \quad (26)$$

- Actions are generated once at a time by the np-rules:

$$occ(AC^i, T) \leftarrow action(AC^i), not\ abocc(AC^i, T) \quad (27)$$
$$abocc(AC^i, T) \leftarrow action(AC^i), action(AC^j),$$
$$occ(AC^j, T), AC^i \neq AC^j \quad (28)$$

- The goal expression $\mathcal{G} = g_1 \wedge \ldots \wedge g_m$ is encoded as
$$goal \leftarrow holds(g_1, T), \ldots, holds(g_m, T) \quad (29)$$

## VI. CORRECTNESS

In this section we prove that the probabilistic answer sets of the np-program encoding of an action theory, **PT**, correspond to trajectories in **PT**, with associated value function. Moreover, we show that the complexity of finding a policy for **PT** in our approach is NP-complete. Let the domain of $T$ be $\{0, \ldots, n\}$. Let be a $T$ transition function associated with **PT**, $s_0$ be a possible initial state, and $a_0, \ldots, a_{n-1}$ be a set of actions in $\mathcal{A}$. Recall, any action $a_i$ can be represented as $a_i = \{a_{1_i}, \ldots, a_{m_i}\}$. Therefore, a trajectory $s_0, \pi(s_0), s_1, \pi(s_1), \ldots, s_{n-1}, \pi(s_{n-1}), s_n$ in **PT** can be also represented as $s_0\ a_{j_0}\ s_1 \ldots a_{j_{n-1}}\ s_n$ for $(1 \leq j \leq m)$ and $(0 \leq i \leq n)$, such that $\forall (0 \leq i \leq n)$, $s_i$ is a state, $a_i$ is an action, $a_{j_i} \in a_i = \{a_{1_i}, \ldots, a_{m_i}\}$, $a_{j_i} = \pi(s_i)$, and $s_i = \Phi(a_{j_{i-1}}, s_{i-1})$.

*Theorem 1:* Let **PT**, be an action theory in $\mathcal{A}_{\mathcal{PO}}$, $\pi$ be a policy in **PT**, and $T_\pi$ be the set of trajectories in $\pi$. Then, $s_0, \pi(s_0), s_1, \pi(s_1), \ldots, s_{n-1}, \pi(s_{n-1}), s_n$ is a trajectory in $T_\pi$ iff $occ(\pi(s_0), 0), \ldots, occ(\pi(s_{n-1}), n-1)$ is true in a probabilistic answer set of $\Pi_{\mathbf{PT}}$.

Intuitively, an action theory, **PT** in $\mathcal{A}_{\mathcal{PO}}$, can be encoded to an np-program, $\Pi_{\mathbf{PT}}$, whose probabilistic answer sets correspond trajectories in **PT**.

*Theorem 2:* Let $h$ be a probabilistic answer set of $\Pi_{\mathbf{PT}}$, $\pi$ be a policy in **PT**, and $T_\pi$ be the set of trajectories in $\pi$. Let $\mathcal{OCC}$ be a set that contains $h \models \tau = occ(\pi(s_0), 0), \ldots, occ(\pi(s_{n-1}), n-1)$ iff $s_0, \pi(s_0), s_1, \pi(s_1), \ldots, s_{n-1}, \pi(s_{n-1}), s_n \in T_\pi$. Then,
$\sum_{h \models value(v,n)\ and\ h \models \tau \in \mathcal{OCC}} v = V_n^\pi(s_0)$

Theorem (2) states that the summation of the values $v$, appearing in $value(v, n)$ that is satisfied by a probabilistic answer set $h$ in which $occ(\pi(s_0), 0), \ldots, occ(\pi(s_{n-1}), n-1)$ is satisfied is equal to the expected sum of discounted rewards after executing a policy $\pi$ starting from a state $s_0$.

The np-program encoding of the reinforcement learning problems, in finite-horizon POMDP, finds optimal policies using the flat representation of the problem domains. Flat representation is the explicit enumeration of world states [23]. Hence, Theorem 4 follows directly from Theorem 3

*Theorem 3 ([23]):* The stationary policy existence problem for finite-horizon POMDP in the flat representation is NP-complete.

*Theorem 4:* The policy existence problem for a reinforcement learning problem in POMDP environment using $NHPLP_{\mathcal{PO}}$ with probabilistic answer set semantics is NP-complete.

## VII. REINFORCEMENT LEARNING USING ANSWER SET PROGRAMMING

Reinforcement learning problems in POMDP can be also encoded as classical normal logic programs with classical answer set semantics [7]. Excluding the np-rules (15), (18), (21) – (24) from the np-program encoding, $\Pi_{\mathbf{PT}}$, of **PT**, results np-program, denoted by $\Pi_{\mathbf{PT}}^{normal}$, with only annotations of the form 1. As shown in [34], the syntax and semantics of this class of np-programs is equivalent to classical normal logic programs with classical answer set semantics.

*Theorem 5:* Let $\Pi_{\mathbf{PT}}^{normal}$ be the *normal logic program* resulting after deleting the np-rules (15), (18), (21) -- (24) from $\Pi_{\mathbf{PT}}$ and $\pi$ be a policy in **PT**. Then, $s_0, \pi(s_0), s_1, \pi(s_1), \ldots, s_{n-1}, \pi(s_{n-1}), s_n$ is a trajectory in $\pi$ iff $occ(\pi(s_0), 0), \ldots, occ(\pi(s_{n-1}), n-1)$ is true in an answer set of $\Pi_{\mathbf{PT}}^{normal}$.

Theorem 5 shows that classical normal logic programs with answer set semantics can be used to solve reinforcement learn-ing problems in POMDP in two steps. First, a reinforcement learning problem, **PT**, is encoded to a classical normal logic program whose answer sets correspond to valid trajectories in **PT**. From the answer sets of the normal logic program encoding of **PT**, we can determine the set of trajectories $T_\pi$ for a policy $\pi$ in **PT**. Second, the value of the policy $\pi$ is calculated using (5). Moreover, any reinforcement learning problem in POMDP environment can be encoded as a SAT problem. Hence, state-of-the-art SAT solvers can be used to solve reinforcement learning problems. Any normal logic program, $\Pi$, can be translated into a SAT formula, S, where the models of S are equivalent to the answer sets of $\Pi$ [19]. Therefore, the normal logic program encoding of a reinforcement learning problem **PT** can be translated into an equivalent SAT formula, where the models of S correspond to valid trajectories in **PT**.

*Theorem 6:* Let **PT** be an action theory and $\Pi_{\mathbf{PT}}^{normal}$ be the normal logic program encoding of **PT**. Then, the models of the SAT encoding of $\Pi_{\mathbf{PT}}^{normal}$ are equivalent to valid trajectories in **PT**.

Reinforcement learning problems can be directly encoded to SAT [32]. This is shown by following corollary.

*Corollary 1:* Let **PT** be an action theory. Then, **PT** can be directly encoded as a SAT formula S where the models of S are equivalent to valid trajectories in **PT**.

VIII. CONCLUSIONS AND RELATED WORK

We described a new high level action language, $\mathcal{A}_{PO}$, that allows the factored representation of POMDP. Moreover, we presented a new reinforcement learning framework by relating reinforcement learning in POMDP to NHPLP. The translation from an action theory representation of a reinforcement learning problem in $\mathcal{A}_{PO}$ into an NHPLP program is based on a similar translation from probabilistic planning into NHPLP [28]. The difference between $\mathcal{A}_{PO}$ and the action languages [1], [2], [5], [11], and [17] is that $\mathcal{A}_{PO}$ is a hight level language and allows the factored specification of POMDP.

The approaches for solving POMDP to find the optimal policies can be categorized into two main approaches; dynamic programming approaches and the search-based approaches (a detailed survey on these approaches can be found in [2]). However, dynamic programming approaches use primitive domain knowledge representation. Moreover, the search-based approaches mainly rely on search heuristics which have limited knowledge representation capabilities to represent and use domain-specific knowledge.

In [22], a logical approach for solving POMDP, for probabilistic contingent planning, has been presented which converts a POMDP specification of a probabilistic contingent planning problem into a stochastic satisfiability problem and solving the stochastic satisfiability problem instead. Our approach is similar in spirit to [22] in the sense that both approaches are logic based approaches. However, it has been shown in [29] that NHPLP is more expressive than stochastic satisfiability from the knowledge representation point of view. In [15], based on first-order logic programs without nonmonotonic negation, a first-order logic representation of MDP has been described. Similar to the first-order representation of MDP in [15], $A_{MD}$ allows objects and relations. However, unlike $A_{PO}$, [15] finds policies in the abstract level. But, NHPLP allows objects and relations. [3] presented a more expressive first-order representation of MDP than [15] that is a probabilistic extension to Reiter's situation calculus. However, it is more complex than [15].

## IX. APPENDIX: EXAMPLE

*Example 2:* The np-program encoding of the tiger domain presented in Example 1 is given by $\Pi = \langle R, \tau \rangle$, where $\tau$ is arbitrary and $R$ consists of the following np-rules, in addition to the np-rules (7), (8), (9), (10), (25), (26), (27), (28):

$action(openL_i) \leftarrow \quad action(openR_i) \leftarrow \quad action(listen_j) \leftarrow$

for $1 \leq i \leq 2$ and $1 \leq j \leq 4$. Properties of the world are described by the fluents $tl$ and $htl$ which are encoded in $R$ by the np-rules

$$fluent(tl) \leftarrow \quad fluent(htl) \leftarrow$$

The set of possible initial states are encoded by the p-rules:

$$holds(tl, 0) \leftarrow not\ holds(\neg tl, 0)$$
$$holds(\neg tl, 0) \leftarrow not\ holds(tl, 0)$$
$$holds(tl, 0) \leftarrow holds(htl, 0)$$
$$holds(\neg tl, 0) \leftarrow holds(\neg htl, 0)$$

The initial probability distribution over the possible initial states is encoded by the p-rules

$$state(0) : 0.5 \leftarrow holds(tl, 0), holds(htl, 0)$$
$$state(0) : 0.5 \leftarrow holds(\neg tl, 0), holds(\neg htl, 0)$$

The executability conditions of actions are encoded by the following p-rules

$exec(openL_i) \leftarrow \quad exec(openR_i) \leftarrow \quad exec(listen_j) \leftarrow$

for $1 \leq i \leq 2$ and $1 \leq j \leq 4$. Effects of the $openL$ action are encoded by the p-rules

$$holds(tl, T+1) \leftarrow occ(openL_1, T), exec(openL_1, T), holds(tl, T)$$
$$holds(\neg tl, T+1) \leftarrow occ(openL_2, T), exec(openL_2, T), holds(\neg tl, T)$$

Effects of the $openR$ action are encoded by the p-rules

$$holds(\neg tl, T+1) \leftarrow occ(openR_1, T), exec(openR_1, T), holds(\neg tl, T)$$
$$holds(tl, T+1) \leftarrow occ(openR_2, T), exec(openR_2, T), holds(tl, T)$$

Effects of the *listen* action are encoded by the p-rules

$$observed(htl, T) \leftarrow occ(listen_1, T), exec(listen_1, T), holds(htl, T)$$
$$observed(htl, T) \leftarrow occ(listen_2, T), exec(listen_2, T), holds(htl, T)$$
$$observed(\neg htl, T) \leftarrow occ(listen_3, T), exec(listen_3, T), holds(\neg htl, T)$$
$$observed(\neg htl, T) \leftarrow occ(listen_4, T), exec(listen_4, T), holds(\neg htl, T)$$
$$holds(tl, T+1) \leftarrow occ(listen_1, T), exec(listen_1, T), observed(htl, T)$$
$$holds(\neg tl, T+1) \leftarrow occ(listen_2, T), exec(listen_2, T), observed(htl, T)$$
$$holds(\neg tl, T+1) \leftarrow occ(listen_3, T), exec(listen_3, T), observed(\neg htl, T)$$
$$holds(tl, T+1) \leftarrow occ(listen_4, T), exec(listen_4, T), observed(\neg htl, T)$$

The probability distribution resulting from executing the *listen* action is given by

$$state(T+1) : 0.85 \times V \leftarrow occ(listen_1, T), exec(listen_1, T), state(T) : V, observed(htl, T), holds(tl, T+1)$$
$$state(T+1) : 0.15 \times V \leftarrow occ(listen_2, T), exec(listen_2, T), state(T) : V, observed(htl, T), holds(\neg tl, T+1)$$
$$state(T+1) : 0.85 \times V \leftarrow occ(listen_3, T), exec(listen_3, T), state(T) : V, observed(\neg htl, T), holds(\neg tl, T+1)$$
$$state(T+1) : 0.15 \times V \leftarrow occ(listen_4, T), exec(listen_4, T), state(T) : V, observed(\neg htl, T), holds(tl, T+1)$$

The rewards received from executing the actions are encoded by

$$reward(-100, T+1) \leftarrow occ(openL_1), exec(openL_1)$$
$$reward(10, T+1) \leftarrow occ(openL_2), exec(openL_2)$$
$$reward(-100, T+1) \leftarrow occ(openR_1), exec(openR_1)$$
$$reward(10, T+1) \leftarrow occ(openR_2), exec(openR_2)$$
$$reward(-1, T+1) \leftarrow occ(listen_1), exec(listen_1)$$
$$reward(-1, T+1) \leftarrow occ(listen_2), exec(listen_2)$$
$$reward(-1, T+1) \leftarrow occ(listen_3), exec(listen_3)$$
$$reward(-1, T+1) \leftarrow occ(listen_4), exec(listen_4)$$

The value function is encoded in $R$ by the np-rules:
$$value(V + \lambda^T * U * -100, T+1) \leftarrow value(V, T),$$
$$factor(\lambda), state(T+1) : U, reward(-100, T+1),$$
$$occ(openL_1, T), exec(openL_1, T),$$
$$holds(tl, T), holds(tl, T+1)$$
$$value(V + \lambda^T * U * 10, T+1) \leftarrow value(V, T)$$
$$factor(\lambda), state(T+1) : U, reward(10, , T+1),$$
$$occ(openL_2, T), exec(openL_2, T),$$
$$holds(\neg tl, T), holds(\neg tl, T+1)$$
$$value(V + \lambda^T * U * -100, T+1) \leftarrow value(V, T),$$
$$factor(\lambda), state(T+1) : U, reward(-100, T+1),$$
$$occ(openR_1, T), exec(openR_1, T),$$
$$holds(\neg tl, T), holds(\neg tl, T+1)$$

$$value(V + \lambda^T * U * 10, T + 1) \leftarrow value(V, T),$$
$$factor(\lambda), state(T + 1) : U, reward(10, T + 1),$$
$$occ(openR_2, T), exec(openR_2, T),$$
$$holds(tl, T), holds(tl, T + 1)$$
$$value(V + \lambda^T * U * -1, T + 1) \leftarrow value(V, T),$$
$$factor(\lambda), state(T + 1) : U, reward(-1, T + 1),$$
$$occ(listen_1, T), exec(listen_1, T),$$
$$observed(htl, T), holds(tl, T + 1)$$
$$value(V + \lambda^T * U * -1, T + 1) \leftarrow value(V, T),$$
$$factor(\lambda), state(T + 1) : U, reward(-1, T + 1),$$
$$occ(listen_2, T), exec(listen_2, T),$$
$$observed(htl, T), holds(\neg tl, T + 1)$$
$$value(V + \lambda^T * U * -1, T + 1) \leftarrow value(V, T),$$
$$factor(\lambda), state(T + 1) : U, reward(-1, T + 1),$$
$$occ(listen_3, T), exec(listen_3, T),$$
$$observed(\neg htl, T), holds(\neg tl, T + 1)$$
$$value(V + \lambda^T * U * -1, T + 1) \leftarrow value(V, T),$$
$$factor(\lambda), state(T + 1) : U, reward(-1, T + 1),$$
$$occ(listen_4, T), exec(listen_4, T),$$
$$observed(\neg htl, T), holds(tl, T + 1)$$